\pdfoutput=1

\documentclass[11pt]{article}

\usepackage[review]{EACL2023}

\usepackage{times}
\usepackage{latexsym}

\usepackage[T1]{fontenc}

\usepackage[utf8]{inputenc}

\usepackage{microtype}

\usepackage{inconsolata}
\usepackage{algorithm}
\usepackage{algorithmic}
\usepackage{bm}
\usepackage{multirow}
\usepackage{color}
\usepackage{amsthm,amsmath,amssymb}
\usepackage{mathrsfs}
\usepackage{graphicx}
\usepackage{verbatim}
\usepackage{enumerate}
\usepackage{array}
\usepackage{xcolor,colortbl}
\definecolor{mydarkgreen}{rgb}{0.0,0.5,0.0}
\makeatletter
\newcommand{\thickhline}{%
    \noalign {\ifnum 0=`}\fi \hrule height 1pt
    \futurelet \reserved@a \@xhline
}

\usepackage{hyperref}
\usepackage{cleveref}
\usepackage{makecell}
\usepackage[utf8]{inputenc}
\usepackage{enumitem}
\title{P-TA: Using Proximal Policy Optimization to Enhance Tabular Data Augmentation via Large Language Models}

\author{Shuo Yang, Chenchen Yuan, Yao Rong, Felix Steinbauer \and Gjergji Kasneci\\
Technical University of Munich, Germany \\
\small{\texttt{\{shuo.yang, chenchen.yuan, yao.rong, felix.steinbauer, gjergji.kasneci\}@tum.de}}\\}

\begin{document}
\maketitle
\begin{abstract}
A multitude of industries depend on accurate and reasonable tabular data augmentation for their business processes. Contemporary methodologies in generating tabular data revolve around utilizing Generative Adversarial Networks (GAN) or fine-tuning Large Language Models (LLM). However, GAN-based approaches are documented to produce samples with common-sense errors attributed to the absence of external knowledge. On the other hand, LLM-based methods exhibit a limited capacity to capture the disparities between synthesized and actual data distribution due to the absence of feedback from a discriminator during training. Furthermore, the decoding of LLM-based generation introduces gradient breakpoints, impeding the backpropagation of loss from a discriminator, thereby complicating the integration of these two approaches. To solve this challenge, we propose using proximal policy optimization (PPO) to apply GANs, guiding LLMs to enhance the probability distribution of tabular features. This approach enables the utilization of LLMs as generators for GANs in synthesizing tabular data. Our experiments demonstrate that PPO leads to an approximately 4\% improvement in the accuracy of models trained on synthetically generated data over state-of-the-art across three real-world datasets. 

\end{abstract}

\section{Introduction}
    With the evolution of business processes that operate on structured data, tabular data has emerged as one of the most crucial data forms, owing to its easily manageable structure and efficient retrievability \citep{gilbert2022analyzing}. Industries such as finance and healthcare heavily rely on tabular data for its clarity and ability to facilitate data analyses and comparisons~\citep{SHWARTZZIV202284}.
    
    However, in realistic applications, the availability of high-quality tabular data is still an issue, mainly due to high costs of data collection and annotation and privacy policies~\citep{10.1145/985692.985752}.
    Therefore, various data augmentation methods have been proposed to overcome the scarcity of high-quality tabular data~\citep{wen2022causaltgan, esmaeilpour2022bi}.

    The contemporary augmentation methods primarily revolve around rules, variational autoencoders (VAE), generative adversarial networks, and LLMs~\cite{wei-zou-2019-eda,Kingma2014, Patki2016TheSD,bao-etal-2019-generating, borisov2023language}. However, these methods exhibit crucial limitations:
    (1) Rule-based methods rely on predefined constraints, which need manual effort and might not capture the diversity of real-world data~\citep{wei-zou-2019-eda}.
    (2) The existing VAE-based and GAN-based approaches primarily focus on matching the distribution of table features rather than whether they are logically coherent~\citep{Kingma2014, Patki2016TheSD}. For example, they might synthesize a sample with an ``\textit{Age}'' of ``\textit{16}'' and an ``\textit{Occupation}'' of ``\textit{Professor}''~\citep{NIPS2014_5ca3e9b1}, resulting in implausible instances due to the lack of external knowledge~\citep{bao-etal-2019-generating}.
    (3) Fine-tuning a LLM on real-world datasets may not capture the real-time differences in the distribution of features between synthetic and real-world data, thereby diminishing the quality of synthesized data. Furthermore, the generation of tabular features via LLMs is implemented through random sampling of the logits of decoders. However, sampling operations are non-differentiable, making it challenging to incorporate GAN-based methods for optimizing a generator, i.e., the LLM, based on the discriminator's output.
    
    To overcome all of these drawbacks, we propose a method of training \underline{P}PO-guided Language models for \underline{T}able \underline{A}ugmentation (P-TA). Specifically, we first transform the tabular data into text using templates. 
    After that, we fine-tune a LLM to generate new data in textual form. 
    Additionally, we train a classifier to distinguish between actual and generated data, utilizing the outputs of the classifier as rewards for further optimizing the LLM through the PPO algorithm~\citep{Schulman2017ProximalPO}.

    Our methodology offers a distinct advantage: PPO-guided LMs can introduce external knowledge, and extensively explore various potential feature combination strategies during training by incorporating GANs, consequently enhancing the likelihood of producing high-quality synthetic data~\citep{sutton1999policy}.
   
    As an additional study for reasonability, we propose to clarify the rationale behind tabular features to enhance the credibility of synthetic data by generating explanations. Explanations can serve as a data auditing tool, specifically for assessing the quality of synthesized data. The data auditing, in turn, contributes to enhancing the accountability of models trained using the synthesized data~\citep{werder2022establishing}. We found the textual explanations are empirically more helpful for data auditors in assessing data quality than plain tables (see user study in our experiments). We are the first to integrate data auditing into tabular augmentation, distinguishing our work from previous efforts~\footnote{The code used in this paper is available at: https://github.com/ShuoYangtum/P-TA.}. 

    Our \textbf{contributions} are as follows:
    \vspace{-8pt}
    \begin{itemize}[itemsep=-5pt]
        \item{We propose P-TA, a novel framework to use PPO-guided language models for tabular augmentation, incorporating GANs to LLMs to generate tabular rows.}
        \item{We benchmark our framework with state-of-the-art (SOTA) methods on three realistic datasets. Training with our synthesized data improves the accuracy by 4\% compared to the SOTA baseline, highlighting the high quality and potential impact of our approach in practical applications.}
        \item{We are the first to explain tabular data via retrieval enhancement. Our user study empirically shows that our explanations successfully assist users with auditing the generated data. }
    \end{itemize}

    \begin{figure*}[t]
        \centering
        \includegraphics[width=0.98\linewidth]{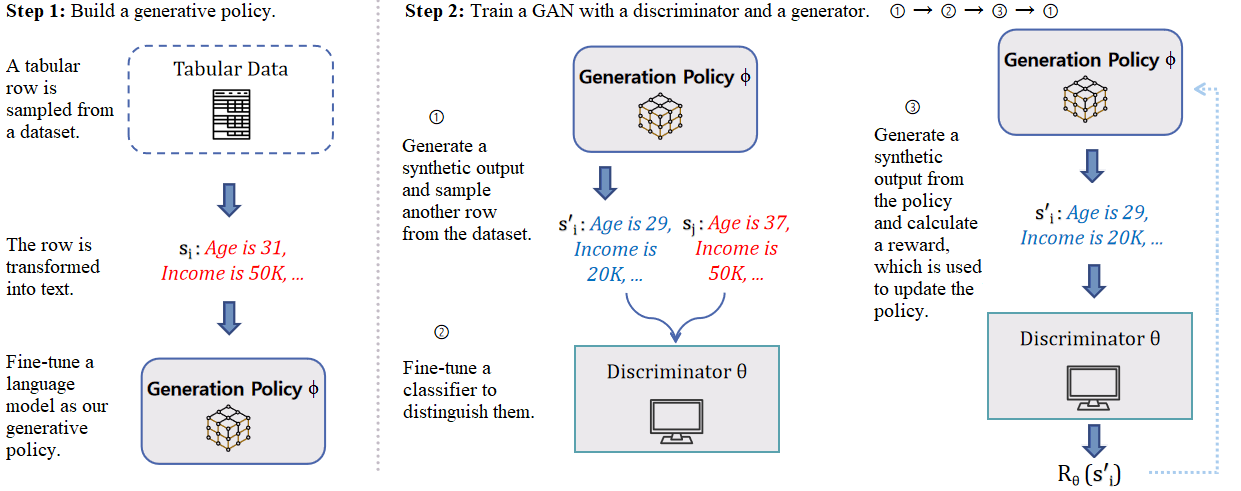}
            \caption{Training comprises two steps: 1) building an initial generative policy, 2) training a discriminator to distinguish the generated samples from ground truth, and updating the generative policy via PPO. }
                \vspace{-15pt}
            \label{fig:ppo}
    \end{figure*}

    \vspace{-5pt}
\section{Related Work}
    With the development of databases, diverse sectors widely use tabular data as an indispensable data form~\citep{gilbert2022analyzing}. 
    However, the accessibility of high-quality tabular data remains a persistent challenge, which is primarily attributed to privacy constraints and the substantial costs in collection~\citep{10.1145/985692.985752}.
    These factors reflect the critical need to propose reasonable tabular synthesizing and auditing technologies.
    \subsection{Tabular augmentation}
        In prior work, table augmentation is mainly based on statistics~\citep{Kamthe2021CopulaFF}, e.g., the Chow-Liu approximation~\citep{1054142}. 
        Contemporary approaches leverage heuristic algorithms, with methods based on VAEs~\citep{10.5555/3495724.3496667, Darabi2021SynthesisingMM} or GANs~\citep{pmlr-v68-choi17a, 10.14778/3231751.3231757, Koivu2020SyntheticMO}.
        As representative approaches, TVAE~\citep{10.5555/3454287.3454946} is optimized based on the evidence lower-bound (ELBO). It maps tabular features into a Gaussian distribution in the latent space and generates new data through sampling; CTGAN~\citep{10.5555/3454287.3454946} adopts a GAN by conditioning the generation process solely on a single discrete feature for tabular data; CopulaGAN~\citep{Kamthe2021CopulaFF},  the current state-of-the-art approach for GAN-based methods, simplifies the underlying CTGAN by utilizing Gaussian copulas.
        However, due to the lack of external knowledge, these models may generate tables with common-sense errors~\citep{9998482}. Consequently, we use LLMs as a knowledge base to introduce a PPO-guided table generator focusing on plausibility and common-sense knowledge. This makes the advantages of LLMs and GANs to be seamlessly integrated.
        
        In the field of generative LLMs, prior to the integration of reinforcement learning, training methods primarily relied on masks~\citep{kenton2019bert, Liu2019RoBERTaAR, lewis-etal-2020-bart} and autoregressive learning~\citep{10.5555/3455716.3455856, NEURIPS2020_1457c0d6}. 
        After that, the most advanced LLMs such as InstructGPT~\citep{ouyang2022training} and the family of LLaMA~\citep{meta2023introducing, alpaca} demonstrated the powerful capabilities of the PPO algorithm mechanism.
    \vspace{-2pt}
    \subsection{Feature interpretation}
        Explainable Artificial Intelligence (XAI) methods have demonstrated their effectiveness in data auditing~\citep{Zhang2017InterpretingCK}.
        Shapley additive explanations (SHAP)~\citet{lundberg2017unified, 10.1145/3307339.3343255} quantify the impact of each feature on the model's output. \citet{ribeiro-etal-2016-trust} simulates complex models by training interpretable models to generate local explanations for specific data points. Gradient-based interpretability models, such as Grad-CAM~\citep{8237336}, IntegratedGrad~\citep{10.5555/3305890.3306024}, generate heatmaps as explanations by computing gradients.
        
        However, these methods emphasize feature attribution concerning predictive impact but do not explain the meaningfulness or plausibility of features in a given context. 
        Therefore, we elucidate the reasons behind inference results, aiming to enhance the credibility of our approach to generate textual explanations. 
    \vspace{-2pt}    
\section{Methodology}
    \vspace{-2pt}
        Given a table with $N$ rows and $M$ columns, where each row represents a sample, and each column represents a feature. 
        We denote this table as $T = (s_1, s_2, ..., s_N)$, where the $i$-th sample row is represented as $s_i = (a_{i1}, a_{i2}, ..., a_{iM})$, with $a_{ij}$ denoting the $j$-th feature column of this sample.

        Regarding data augmentation, our target is to generate a new table, denoted as $T' = (s_1', s_2', ..., s_K')$, which contains $K$ synthetic samples. 
        The synthetic dataset should exhibit a distribution that resembles the original dataset $T$.
        
         As for data auditing, our target is generating textual explanations for the plausibility of a given feature value $a_{ij}'$ in an arbitrary new sample $s_i'$. 
    \vspace{-4pt}
    \subsection{Tabular data augmentation}
    \label{sec:tda}
        In the first step, we expand the tabular data by converting it into text through a predefined template. We then employ a LLM to generate new textual samples. 
        Finally, we transform the generated text into tabular form via the same template. 
        \Cref{fig:ppo} demonstrates the training pipeline.

    \paragraph{Transforming tabular data into textual data.} 
        We employed an effective transformation template of ``[Feature] is [Value]''~\cite{borisov2023language,zhang2023generative}.
        Here, [Feature] represents a specific feature name in a sample, and [Value] denotes the corresponding value. 
        By connecting short phrases associated with all features using commas, we construct a sentence describing a sample. 

    \paragraph{Training} 
        We initiate fine-tuning a LLM $\phi$ on these sentences. 
        The LLM is treated as the generator of a GAN and a knowledge base to mitigate the logical and semantic inconsistencies or conflicts among the generated features~\citep{heinzerling-inui-2021-language}. Our objective is to enable the LLM to emulate the underlying distribution patterns of the feature values and minimize the generative perplexity, as shown in Eq.~\eqref{eq:LM_loss}:
            \begin{equation}
                \label{eq:LM_loss}
                \mathcal{L}_\text{LLM}(\phi)=-\sum_{i} \log p(w^{\text{LLM}}_i=w_i|c_i;\phi),
            \end{equation}
           where $c_i$ is the context of a target word $w_i$. $w^{\text{LLM}}_i$ is the corresponding prediction of the LLM.
   
       We then generate new samples by sampling from the logits of LLM $\phi$. 
       The probability of generating the $i$-th token $w_i$ as $w\in V$ is then given by $P_{\phi}(w|w_{1:i})$, where $V$ represents the vocabulary. 
       To encourage the LLM to generate diverse samples, we utilize a temperature function~\citep{ACKLEY1985147} with a small $\tau$:
        \begin{equation}
            P_{\phi}'(w_i)=\frac{P_{\phi}(w|w_{1:i})^{1/\tau}}{\sum_{w' \in V}P_{\phi}(w'|w_{1:i})^{1/\tau}}.
        \end{equation}
        To prevent text degeneration, we employ the top-$p$ sampling~\citep{Holtzman2020The} and sample token $w_i$ according to probability distribution $P_\phi''$:
        \begin{equation}
            P_{\phi}''(w_i) = \begin{cases}
            P_{\phi}'(w_i)/\sum_{w' \in V_p}P'_{\phi}(w') &\text{if }w_i \in V_p\\
            0 & \text{otherwise},
            \end{cases}
        \end{equation}
        where top-$p$ vocabulary $V_p$ is the smallest set such that: $\sum_{w_i \in v_p} P(w_i|w_{1:i-1}) \geq p$.
    
        Finally, we repeat the sampling and utilize the mentioned template to transform generated sentences back into tabular rows to obtain $T'$.

        However, the missing or complex irregular dependencies between attributes in tables~\citep{9998482} can lead to model biases and reduced robustness. 
        Since GAN-based methods have proven to excel in generating text with given features~\citep{yu2017seqgan}, we employ the PPO algorithm to optimize the performance of the fine-tuned LLM~\citep{ouyang2022training} by incorporating it as the generator within a GAN.

        In particular, we train a classifier $\theta$ to distinguish between synthetic data labeled as $y^\text{s}$ and real data labeled as $y^\text{r}$. Here, $\theta$ is trained using a focal loss~\citep{8417976} to mitigate potential issues related to class imbalance.
        Then, we calculate the reward of a generated sample $s'$ via the classifier $\theta$:
        \begin{equation}
           \text{R}_\theta(s')=P(y'=y^\text{r}|s';\theta),
        \end{equation}
        where the reward value is the probability that $y'$ is classified as actual data by the classifier $\theta$. 


        We then employ PPO to increase the probability of generating sentences with high reward values, which theoretically leads to a closer alignment between generated and original data distributions. 
        The objective function for model output $s'$ consists of two components: 
        the score $R_\theta(s')$ computed by the reward model and the KL divergence between the generative policy and the sampling policy, i.e., an orininal copy of the generative policy~\citep{ouyang2022training}. Hence, the goal is to maximize:
        \begin{equation}
            \begin{aligned}
                E_{s' \in T'}[R_\theta(s') 
                -\beta \log(P^\text{RL}(s';\phi)/ P^\text{SFT}(s'))],
            \end{aligned}
        \label{eq:ppo}
        \end{equation}
        $\beta$ is the weight coefficient, $P^\text{RL}(s')$ and $P^\text{SFT}(s')$ are the probability related to the generative policy and the sampling policy for $s'$, respectively. 

        Here, we have two training objectives for the generator: (1) maximizing reward values by aligning distributions of synthetic data and actual data, and (2) minimizing the Kullback–Leibler divergence between the generation policy and the sampling policy.

        Finally, we iterate through the training process until the GAN, i.e., the generator $\phi$ and the classifier $\theta$, reaches a Nash equilibrium~\citep{doi:10.1073/pnas.36.1.48}.

    \subsection{Data auditing}
         \begin{algorithm}[ht]
        \caption{Feature Interpretation Algorithm.}
        \label{al:inter}
        \begin{algorithmic}[1]
           \STATE {\bfseries Input:} A generated tabular row $s'=(a_{1}', a_{2}', ..., a_{M}' )$, with the $j$-th feature value $a_{j}'$ to be explained;
                The original dataset $T=(s_1, s_2, ..., s_N)$.
           \STATE {\bfseries Output:} A textual explanation for why $a_{ij}$ is reasonable in context $s_i'$.
            \STATE {\small\color{lightgray} \# Generate textual descriptions.}
            \STATE Use prompt learning to transform $s'$ and all $s_i \in T$ into their textual descriptions $d'$ and $(d_1, d_2, ..., d_N)$.
            \STATE {\small\color{lightgray} \# Retrieve descriptions using the augmented strategy.}
            \STATE Retrieve $k$ most similar descriptions $(d_{\hat{1}}, ..., d_{\hat{k}})$ to $d'$ from $(d_1, d_2, ..., d_N)$.
            \STATE {\small\color{lightgray} \# Generate explanations based on descriptions.}
            \STATE Generate explanation using $(d_{\hat{1}}, ..., d_{\hat{k}})$ and $d'$.
        \end{algorithmic}
    \end{algorithm}
        \vspace{-5pt}
        
        We aim to enhance the plausibility of our model by explaining and providing reasons for the generated feature values. On the one hand, reading these explanations can enhance the transparency of table synthesis techniques for users. On the other hand, when this technology is applied to real-world data, it allows table users to assess the quality of tables without direct access to the table contents, thus ensuring data privacy in theory. We first describe each tabular row as a paragraph that rephrases all its features into text via a LLM. Then, the interpreter reads and compares these textual descriptions of the original samples to explain the reasons behind the presence of a particular value for a given feature. Algorithm~\ref{al:inter} presents the steps towards such a plausibility-related interpretation.

                    \begin{table*}[ht]
    \resizebox{\textwidth}{!}{
    \begin{tabular}{ll}
    \thickhline
    Travel Customers & \url{https://www.kaggle.com/datasets/tejashvi14/tour-travels-customer-churn-prediction} \\
    Adult Income               & \url{https://archive.ics.uci.edu/dataset/2/adult} \\
    HELOC               & \url{https://kaggle.com/datasets/averkiyoliabev/home-equity-line-of-creditheloc} \\ \thickhline
    \end{tabular}}
    \caption{The real-world datasets used in our study.}
    \label{tab:dataset}
    \end{table*}
    
        \paragraph{Generation of textual descriptions.}
        We first convert a table row into a sentence using the template described in Section~\ref{sec:tda}, then fill it into a prompt template, e.g., ``Please describe a person with the following features.'' Following our instructions, the interpreter pre-processes the templates into descriptions before generating the explanation.
        
        Pre-processing tabular rows to descriptions for explanation holds three theoretical advantages: First, the LLM naturally incorporates additional information. 
        For instance, for an individual with the job title ``\textit{professor}'' age ranges between 0 and 25 years are less plausible, while a ``\textit{busy schedule}'' may be quite likely. 
        Therefore, these supplementary details can aid the interpreter in inferring more profound and intricate underlying reasons. Secondly, the tabular features always contain numerous abbreviations, many needing to be more readily understandable by language models and individuals who need domain expertise. By using a LLM to convert them into text descriptions, they should be more comprehensible. Thirdly,  \cite{10.1145/3331184.3331303} has demonstrated that using descriptive text outperforms using keywords (such as tabular feature values) in information retrieval tasks, which serves as empirical support for our discussion in the next paragraph.
            \begin{table*}[ht]
    \centering
    \resizebox{\textwidth}{!}{
    \begin{tabular}{lccccccccc}
    \thickhline
    \multicolumn{2}{c}{Dataset}                              & \multicolumn{1}{c}{Original} & \multicolumn{1}{c}{TVAE} & \multicolumn{1}{c}{CopulaGAN} & \multicolumn{1}{c}{CTGAN} & \multicolumn{1}{c}{Neo + GReaT} & \multicolumn{1}{c}{GPT-2 + GReaT} & \multicolumn{1}{c}{Neo + Ours} & \multicolumn{1}{c}{GPT-2 + Ours} \\ \hline
    \multirow{4}{*}{Travel}   & LR $\uparrow$                      & 84.29\% & 80.64\% & 74.23\%                       & 75.12\%  & 74.82\%  & 79.11\%                          & \underline{80.95\%}                             & \textbf{82.22\%}                  \\
              & DT $\uparrow$   & 86.91\%  & \textbf{80.90\%}  & 75.61\%  & 73.30\%      & 72.73\%  & 74.68\% & 74.81\%  & \underline{80.14\%}   \\
             & RF $\uparrow$  & 87.43\%   & \underline{80.92\%}   & 74.23\%  & 72.23\%    & 75.04\%    & 75.31\%    & 69.05\%        & \textbf{81.53\%}      \\
              & \multicolumn{1}{l}{Mean} & 86.21\%         & \underline{81.01\%}  & 74.69\%    & 73.55\%    & 74.19\%  & 76.37\%  & 74.94\%   & \textbf{81.30\%}   \\ \hline
    \multirow{4}{*}{Adult Income } & LR $\uparrow$                      & 80.55\%   & \underline{77.45\%}                    & 76.79\%  & \textbf{79.58\%}            & 76.49\%   & 75.69\%   & 76.56\%     & 76.27\% \\
         & DT $\uparrow$  & 82.36\%       & \underline{79.60\%}   & 72.44\%   & 78.50\%   & 73.10\%   & 73.17\%  & \textbf{79.67\%}           & 73.27\%   \\
            & RF $\uparrow$    & 85.81\%       & 79.60\%  & 77.46\%   & 79.63\%  & \underline{80.38\%}  & 79.79\%  & \textbf{80.44\%}   & 79.82\%  \\
            &  Mean   & 82.91\%      & 78.88\%     & 75.56\% & \textbf{79.24\%}   & 76.66\%   & 76.22\% & \underline{78.89\%}  & 76.45\%   \\ \hline
    \multirow{4}{*}{HELOC}        & LR $\uparrow$         & 69.79\%    & 61.04\%   & 42.03\%  & 57.72\%      & \underline{64.39\%}     & 52.44\%  & \textbf{75.65\%} & 51.64\%   \\
          & DT $\uparrow$  & 62.00\% & \textbf{66.39\%}   & 42.36\%     & 61.34\%  & \underline{60.16\%}   & 56.51\%   & 55.03\%    & 57.14\%  \\
         & RF $\uparrow$   & 70.12\% & 67.24\%              & 42.35\%      & 62.35\%  & \underline{71.20\%}   & 62.80\%   & \textbf{77.83\%}   & 66.70\% \\
        &  Mean  & 67.30\%    & 64.89\%  & 42.25\%   & 60.47\%     & \underline{65.25\%}    & 57.25\%    & \textbf{69.50\%} & 58.49\%                          \\ \thickhline 
        \end{tabular}
        }
        \caption{Accuracy measure. LR, DT, and RF stand for Logistic Regression, Decision Tree, and Random Forest, respectively. \textbf{Bold} indicates the best performance, and \underline{underline} indicates the second best in terms of being closest to the accuracy on the original data.}
        \label{tab:accuracy}
        \end{table*}
        
        \paragraph{Retrieval augmented generation.}
        When explaining a generated feature value, we retrieve similar samples corresponding to a target sample as comparative data to our interpreter, an LLM. The advantage of retrieval lies in providing the interpreter model with comparative information, thereby preventing incorrect or incomplete analysis. For instance, if two individuals who are otherwise identical except for age receive different levels of ``income'', the interpreter can theoretically infer that age is the most significant factor causing the disparity in the ``income'' feature. 

        To retrieve similar samples, we compare two samples by comparing the semantic similarity of their descriptive texts.
        Specifically, we employ a LLM that has been fine-tuned on semantic similarity tasks to embed the descriptive texts corresponding to two samples separately. 
        After that, we calculate the cosine similarity between the embeddings as a measure of sample similarity.
        Finally, the top $k$ most similar samples retrieved are input into the interpreter for explanation.

    \begin{table*}[ht]
    \centering
    \setlength{\abovecaptionskip}{3pt}
    \resizebox{\textwidth}{!}{
    \begin{tabular}{lrrrcccc}
    \thickhline
                 & \multicolumn{1}{c}{CopulaGAN} & \multicolumn{1}{c}{CTGAN} & \multicolumn{1}{c}{TVAE} & \multicolumn{1}{c}{Neo + GReaT} & \multicolumn{1}{c}{GPT-2 + GReaT} & \multicolumn{1}{c}{Neo + Ours} & \multicolumn{1}{c}{GPT-2 + Ours} \\ \hline
    Travel $\downarrow$      & 97.58\%                         & \underline{92.13\%}    & \textbf{91.26\%}    & 98.05\%   & 98.04\%  & 96.15\%    & 96.26\%   \\
    Adult Income $\downarrow$ & 88.47\%                         & 87.65\%     & 96.92\%                    & 82.98\%     & \underline{82.73\%}  & \textbf{82.51\%}   & \underline{82.73\%}                     \\
    HELOC $\downarrow$   & 98.61\%    & 100.00\%   & 99.97\%     & 92.79\%     & 91.50\%    & \textbf{86.80\%}  & \underline{89.13\%}                     \\ 
    \rowcolor{lightgray} Mean $\downarrow$     & 94.89\%    & 93.26\%    & 95.85\%    & 91.27\%    & 90.76\%     & \textbf{88.49\%}     & \underline{89.37\%}                     \\ \thickhline
    \end{tabular}}
    \caption{Discriminator measure. We use decreased accuracy values to check whether the data generated cannot be easily distinguished apart from the original data.}
    \label{tab:dis}
    \end{table*}

\section{Experiments} 

    \subsection{Datasets}
    We use three real-world datasets for evaluation by following \citep{borisov2023language}. The data we utilize can be directly downloaded from the internet, as shown in Table~\ref{tab:dataset}. The dataset \textbf{Travel} encompasses information about travel customers aimed at aiding a travel company in computing its churn rate. We utilize the ``\textit{Target}'' feature, denoting customer attrition, as both the classification label and the feature to be audited.
    \textbf{Adult Income} consists of records from the 1994 Census database. 
    It comprises 48,842 instances and 14 attributes. We treated the ``\textit{Income}'' feature, denoting the annual income of specific individuals, as both the classification label and the feature to be audited.
    \textbf{HELOC} dataset comprises 10,460 samples, each with 24 features related to anonymized information about \emph{home equity line of credit} (HELOC) applications made by real homeowners. 
    We regard the ``RiskPerformance'' feature as the classification label and the feature to be audited.

    For evaluation, we generate 10,000 synthetic samples from Adult Income and HELOC and 1,000 synthetic samples from Travel Customers due to the relatively low data size and feature count in the latter dataset.

        \subsection{Automatic metrics}
        \label{sec:automatic}
        \paragraph{Accuracy.} 
        Following prior work~\citep{borisov2023language}, we first conduct training using the original data on three classifiers: logistic regression (LR)~\citep{abd18569-0124-3b1e-9ca9-67a6fd857a26}, decision trees (DT)~\citep{breiman1984classification}, and random forests (RF)~\citep{liaw2002classification}. The training involve inputting the feature vectors of original samples into these classifiers and making predictions of the target feature. 
        Subsequently, we supply the feature vectors of the generated samples as test samples to these classifiers and observe whether the predicted values by the classifiers match the generated values of the target variable. 
        The closer the predictive accuracy remains to the original accuracy, i.e., accuracy on the original dataset, the more similar the distribution of the test dataset, i.e., the generated data, is to the distribution of the original dataset. 
        \vspace{-5pt}
        \paragraph{Discriminator measure.} We use the accuracy of a classifier trained to distinguish between original and synthetic data to measure the performance of our generation approach. 
        High-quality synthetic data should render the trained classifier incapable of accurately categorizing whether the data is synthetic or not. 
        
        To mitigate information leakage, we use a Kernel Support Vector Machine~\citep{svm}, contrasting to the neural network-based classifier used in training. 
        \paragraph{Jaccard Coefficient.} It is a statistical metric employed to quantify the similarity between two sets~\citep{niwattanakul2013using}. We compute the Jaccard coefficient between the feature sets of a synthetic sample and the sample in the original dataset with the maximum feature overlap for that synthetic sample. A larger Jaccard coefficient indicates a closer similarity between the feature value distributions of synthetic data and the original data.

    \subsection{Human evaluation}
    \label{sec:manual}
    We conduct data audits with human subjects by presenting them explanations generated by our model. Feedback from the pilot study indicates that auditing the HELOC dataset requires strong financial knowledge. Therefore, we conduct data audits on two datasets, Adult Income and Travel, for high-quality human evaluation.  
    On each dataset, ten randomly selected samples are shown. For explanation quality assessment, human users rate the quality regarding \textbf{reasonableness, relevance, clarity, and comprehensiveness} within a 7-point Likert scale~\citep{likert1932technique}. Contrastively, we apply SHAP to generate numerical explanations. We asked all participants to indicate their preference for encountering either textual or quantified numerical explanations.

    \begin{table*}[ht]
    \centering
    \resizebox{\textwidth}{!}{
    \begin{tabular}{ccccccccccccc}
    \thickhline
            &  & \multicolumn{5}{c}{Ours}         &  & \multicolumn{5}{c}{SHAP}         \\ \cline{3-7} \cline{9-13} 
    Dataset &  & REA. & COM. & CLA. & REL. & Mean &  & REA. & COM. & CLA. & REL. & Mean \\
    Travel  &  & 4.29$\pm$0.19 & 4.32$\pm$0.19 & 4.87$\pm$0.18 & 4.37$\pm$0.20 & 4.46 &  & 3.69$\pm$0.21 & 3.93$\pm$0.21 & 3.11$\pm$0.24 & 4.26$\pm$0.22 & 3.75 \\
    Adult   &  & 4.26$\pm$0.20 & 4.00$\pm$0.22 & 5.10$\pm$0.17 & 4.22$\pm$0.20 & 4.40 &  & 3.64$\pm$0.20 & 4.17$\pm$0.21 & 3.14$\pm$0.23 & 3.74$\pm$0.20 & 3.67 \\
    \rowcolor{lightgray} Mean    &  & \textbf{4.28} & \textbf{4.27} & \textbf{4.99} & 4.19 & \textbf{4.43} &  & 3.67 & 3.84 & 3.13 & \textbf{4.22} & 3.71 \\ \thickhline
    \end{tabular}}
        \caption{Human auditing results with 0.95 confidence interval, where ``REA.'' represents reasonableness, ``COM.'' represents comprehensive, ``CLA.'' represents clarity, ``REL.'' represents relevance.}
        \label{tab:human}
        \vspace{-5pt}
    \end{table*}

        \begin{figure*}[t]
        \centering
       \includegraphics[width=0.90\linewidth]{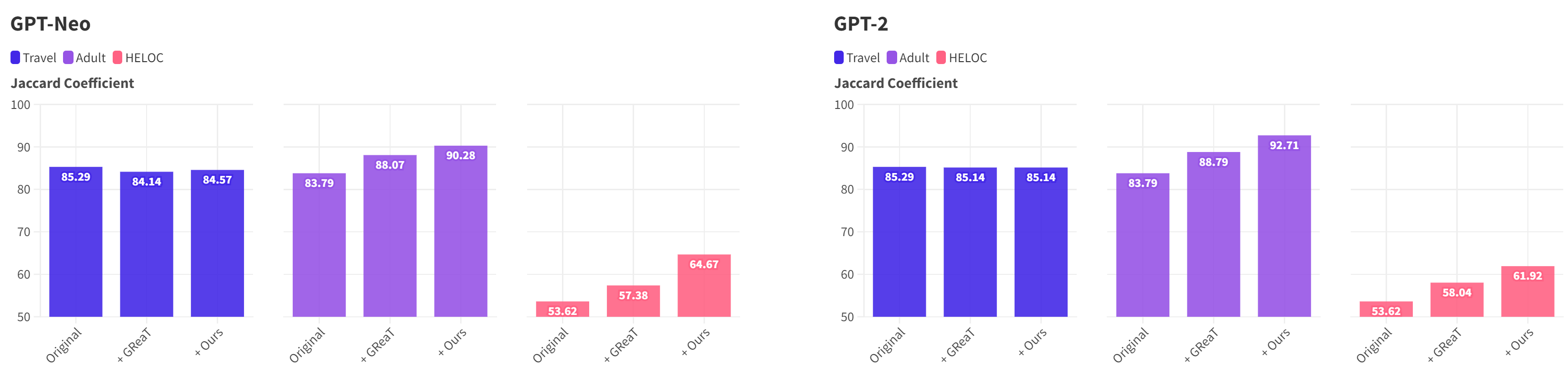}
            \caption{Jaccard Coefficient between generated and original samples in the training set. (Scaled by the factor of $100$ for a clear view.) }
            \label{fig:Jaccard}
            \vspace{-5pt}
    \end{figure*}

    \begin{table*}[ht]
    \centering
    \setlength{\abovecaptionskip}{3pt}
    \resizebox{\textwidth}{!}{
    \begin{tabular}{p{1.5cm} p{2cm} p{3.5cm} p{9.5cm}}
    \thickhline
    \multicolumn{2}{l}{\multirow{2}{*}{Travel}} & \small{Synthetic data}         & \small{age:49, workclass:Private, ... education:HS-grad, educational-num:9, marital-status:Married-civ-spouse, occupation:Craft-repair, ..., income:>50K} \\
    \multicolumn{2}{l}{User}                            & \small{User's question}        & \small{Explain the reason why the person has an income over 50K.} \\ \hline
    \multicolumn{2}{l}{Ours}    &     \small{Explanation }                   & \small{... his \textcolor{blue}{educational background} may have contributed to his ability to negotiate better pay and benefits packages ... given \textcolor{blue}{his marriage to someone with similar socioeconomic characteristics}, such as being HS-grad and married ... increasing their joint income potential ...}  \\ \thickhline
    \end{tabular}
    }
    \caption{Data audit using our explanation on Adult Income Dataset. \textcolor{blue}{Blue text} represents the main reason in the explanation. Please refer to \cref{sec:appendix-casestudy} for the full explanation.
    }
    \label{tab:case_travel}
    \end{table*}
    
       \begin{table*}[ht]
    \setlength{\abovecaptionskip}{3pt}
       \centering
       \resizebox{\textwidth}{!}{
    \begin{tabular}{p{1.5cm} p{2cm} p{3.5cm} p{9.5cm}}
    \thickhline
    \multicolumn{2}{l}{\multirow{2}{*}{Travel}} & \small{Synthetic data}         & \small{Age:30, ..., IncomeClass:Middle Income, ServicesOpted:2, AccountSyncedToSocialMedia:No, BookedHotel:Yes} \\
    \multicolumn{2}{l}{User}                            & \small{User's question}        & \small{Explain the reason why the person hasn't churned.} \\ \hline
    \multicolumn{2}{l}{Baseline}          & \small{w/o. similar samples retrieved }               &\small{... because they have \textcolor{blue}{booked a hotel}, which indicates that they are still interested in the services offered ... \textcolor{red}{their age falls within the middle income class ... It also does not appear that this customer is a frequent flyer or syncs their account to social media, both of which could potentially indicate higher levels of engagement with the company.}}\\
    \multicolumn{2}{l}{Ours}                            & \small{with similar samples retrieved} & \small{... because \textcolor{blue}{they have engaged positively by choosing two services} ... which indicates some level of satisfaction ... despite not syncing their account with social media platforms, they have still managed to \textcolor{blue}{book a hotel} ... indicating that the company's offerings meet their needs satisfactorily.}  \\ \thickhline
    \end{tabular}
    }
    \caption{Comparison of generated explanation on the Travel Dataset. \textcolor{blue}{Blue text} represents key-points in explanations. \textcolor{red}{Red text} represents unreasonable explanations. Please refer to appendix~\ref{sec:appendix-casestudy} for more details.}
    \label{tab:case_travel}
    \end{table*}
\section{Results}
    \subsection{Analysis for automatic evaluation}
    
        Table~\ref{tab:accuracy} illustrates the accuracy tests conducted on data generated by our approach, where our model outperforms baselines across three distinct datasets. 
        Specifically, on the Adult Income dataset, our model achieves a considerable increase (of 4 percent points) in terms of accuracy, clearly surpassing the state-of-the-art methods. 
        Similarly, our approach surpasses the baseline by $2\%$ points on the HELOC dataset. 
        In the case of the smaller-scale Travel dataset, our PPO techniques lead to a remarkable increase of $5\%$ points. However, the performance of all the models on the HELOC dataset is lower. 
        This should be attributed to the relatively high number of features in HELOC, not all correlated with the target feature being predicted. A substantial amount of redundant information from the multitude of features can potentially interfere with the predictions of these models.

        In Table~\ref{tab:dis}, it is observed that our method outperforms the baselines in terms of decreasing a discriminator's ability to discriminate between real and synthetic samples across all three datasets. Even in comparison to the best performing baseline (GPT-2 + GReaT), a reduction of over $2\%$ points of the mean discrimination measure can be observed. In comparison with CopulaGAN or TVAE, the reduction is even more impressive (with over $6\%$ and $7\%$ points, respectively).  This is an important 689 481 3419advantage of our method, especially for practical applications, where a high reliability of synthesizers is needed.
        We attribute this to the close resemblance of the data generated by our approach to the original data distribution, rendering it challenging for the discriminator to accurately distinguish between them. 
        However, our method performs less effectively than TVAE on the Travel dataset. We hypothesize  that when the number of tabular features is limited, TVAE exhibits similar or more effective data-fitting capabilities than LLMs.

    \subsection{Analysis for human evaluation}
    In the user study, we recruited 30 participants using online recruitment platform Prolific~\footnote{https://www.prolific.com/}. We required the participants to be fluent in English. Each participant was compensated with a payment of £6 for participation in the user study (within 40 minutes). 
    As shown in Table~\ref{tab:human}, we observe that the explanations generated by our system receive positive evaluations (scores $>$ 3.5) across all evaluation dimensions for both datasets. Compared to numerical SHAP explanations, our explanations achieve higher scores in clarity (CLA), reasonableness (REA) and comprehensiveness (COM), which contributes to
    a 19.4\% improvement in the mean score. These results indicate the high quality of our explanations.
    Notably, 80\% and 77.33\% of individuals prefer our explanations for auditing the data to numerical explanations of SHAP in the Adult and Travel datasets, respectively. 

    
     When examining the REA score on each sample, there are samples that are not reasonable. For instance, in the Adult dataset, the example presented in Table~\ref{tab:case_travel} receives the lowest REA score. The synthetic data in this case indicates that corresponding adults have an income of ''$\ge$50K.'' However, all three binary classifiers trained for accuracy classify this example as ''$<$50K,'' indicating that the example is an erroneous synthetic sample. Our users quickly identify logical errors by reading the explanation we provide. 

    \begin{table}[H]
    \centering
    \begin{tabular}{ccccc}
    \thickhline
        & REA  & COM  & CLA  & REL \\ \hline
    SVM. & 0.64 & 0.63 & 0.49 & 0.5 \\ \thickhline
    \end{tabular}
    \caption{Pearson correlation coefficient between human and automatic evaluation by an SVM model. All results yield $p$-value $<0.05$.}
    \label{tab:pearson}
    \vspace{-5pt}
    \end{table}
    
     To confirm that humans can effectively utilize our explanations to identify flaws in generated data that could impact classifier training, we calculate Pearson correlation coefficients among human evaluation metrics to the automatic evaluation by an SVM, which is the probability of a given sample being classified to that class. From Table.~\ref{tab:pearson}, we observe positive correlations, indicating that human and automatic evaluation of data quality exhibit consistency. This demonstrates that, by reading explanations generated by LLMs, people can assess the quality of tables without direct access to the contents, effectively preserving privacy of critical, person-related information. Notably, a robust correlation is found between the REA and ACC metrics, highlighting the power of our explanations.    
    
    \subsection{Analysis for ablation experiment}
    \label{sec:ab}

        In the augmentation task, as depicted in Fig.~\ref{fig:Jaccard}, we observe that PPO can further enhance the similarity of generated data to the original data distribution. 
        This optimization benefits training on relatively large datasets, for instance, on the Adult and HELOC datasets, it improves the Jaccard similarity coefficient by 4\% and 7\%, respectively
        Additionally, we observe consistency in the performance of the two models in Fig.~\ref{fig:Jaccard}. Specifically, GPT-2 achieves higher performance on the Travel dataset, while GPT-Neo exhibites higher scores on the Adult and HELOC datasets. We hypothesize that this might be attributed to GPT-Neo's superior aptitude in handling longer tabular texts. The diversity in the training data for GPT-Neo may enhance its capability to capture textual features in longer documents.

        \Cref{tab:case_travel} demonstrates the advantage of our proposed retrieval strategy. In this example, due to the absence of comparative information, the baseline confuses age and income in assessing the impact on user retention. Furthermore, we observe logical errors in explanations, such as treating "not a frequent flyer" as a positive factor. In contrast, our model analyzes all potential factors. 
\section{Limitation}
    During the training phase, as PPO involves the concurrent utilization of a discriminator and two policies, the time complexity for a single epoch is $2*O(k_1n) + O(k_2n)$, where $k_1$ and $k_2$ represent the time complexity of the generation policy and discriminator in training, respectively, and $n$ denotes the number of features in the tabular data. In contrast, the time complexity for baselines generated based on language models is $O(k_1n)$. In inference, our generation policy shares the same linear time complexity as the table generator GReaT, which is based on a language model. Utilizing a single A100 GPU, generating 1,000 HELOC tabular samples, each consisting of 24 features, incurs an average per-sample generation time of only 19.31 milliseconds. In summary, despite our model exhibiting higher training time complexity, it adequately meets the requirements of real-world applications.
\section{Conclusion}
        Our research tackles critical challenges in tabular data usage, focusing on integrating language models into generative adversarial networks for data augmentation. Specifically, 1)  Proximal Policy Optimization can leverage a discriminator's outputs to optimize a language model's generation strategy, thereby effectively leading to a substantial boost in the plausibility of generated data. 2) We successfully generated explanations for synthetic tabular features to assist data audits. Human evaluations highlight the clarity and the power of our explanations generated for data audits. 
        These contributions could transform how industries obtain and audit tabular data, facilitating more informed decision-making and refined data-driven processes.

\bibliography{anthology,custom}
\bibliographystyle{acl_natbib}

\appendix
    
\section{Ethics Statement}

    In the human evaluation phase, the individuals participating in the evaluation are anonymous to the authors of this paper. 
    We pay them with the local standards and legal requirements. 
    Furthermore, participants will only receive information about the data generated and the research questions mentioned above. We ensured that all participants comprehended the content of the task and agreed to participate in the manual evaluation.

    Regarding model usage, we have obtained official approval from Meta to use LLaMA 2. Additionally, the Vicuna-1.5 model is an open-source supplementary model built upon the existing weights of LLaMA 2.
    
 \section{Additional Experiments and Results}
 We employ the following metrics for implement additional experiments:

    \paragraph{Average negative log-likelihood metric.}
    The generated data is expected to conform to the distribution of the training data. Following~\citep{borisov2023language}, we calculated the Log-likelihood of synthetic data samples on a density model derived from the original data ($L_{syn}$) and of the original test data on the model derived from the synthetic data ($L_{test}$) using a Gaussian Mixture Model for 10,000 samples generated on the Adult Income and HELOC datasets. As depicted in Table~\ref{tab:nll}, our method demonstrates comparability with the state of the art (SOTA). Furthermore, we observed superior performance beyond the baseline concerning $L_{syn}$, although our approach exhibited slightly inferior performance compared to the baseline in terms of $L_{test}$. Our analysis indicates that the PPO algorithm encounters challenges associated with overfitting in more prevalent data modes. Nevertheless, the adverse effects induced by PPO are relatively constrained, given its potential for a more pronounced enhancement in other utilized metrics. 
    
    \begin{table}[ht]
    \resizebox{\linewidth}{!}{
    \begin{tabular}{llrrlrr}
    \hline
              &  & \multicolumn{2}{c}{Adult}                                      &                      & \multicolumn{2}{c}{HELOC}                                      \\ \hline
    Models    &  & \multicolumn{1}{c}{$L_{syn}$} & \multicolumn{1}{c}{$L_{test}$} &                      & \multicolumn{1}{c}{$L_{syn}$} & \multicolumn{1}{c}{$L_{test}$} \\ \cline{3-4} \cline{6-7} 
    Identity  &  & -13.852                       & -13.852                        & \multicolumn{1}{r}{} & -55.672                       & -55.672                        \\
    CTGAN     &  & -11.221                       & \textbf{-21.822}               &                      & -62.584                       & \textbf{-97.382}               \\
    GReaT-2   &  & -10.706                       & -23.684                        &                      & -61.600                       & -127.006                       \\
    GReaT-Neo &  & -10.656                       & -26.054                        &                      & -60.363                       & -134.411                       \\
    Ours-2    &  & -10.630                       & -24.095                        &                      & -59.126                       & -135.449                       \\
    Ours-Neo  &  & \textbf{-10.484}              & -28.433                        & \multicolumn{1}{r}{} & \textbf{-58.260}              & -147.703                       \\ \hline
    \end{tabular}}
    \caption{Average log-likelihood of synthetic and original data. We calculate them by using density models obtained from the original data ($L_{syn}$) and the synthetic data ($L_{test}$), respectively.}
    \label{tab:nll}
    \end{table}

     \paragraph{KL divergence.} It quantifies the difference between two probability distributions~\citep{kullback1951information}. As most of the feature values in tabular data are not of numeric type and have relatively small value ranges, making visualization challenging, we select to visualize and calculate the KL divergence only for the distributions of ``education years'', ``weekly working hours'', and ``age'' in the Adult Income dataset.
     
     Regarding consistency of our human evaluation, we computed Fleiss's Kappa for the scorings of reasonableness, relevance, clarity, and comprehensiveness, resulting in values of 0.16, 0.16, 0.21, and 0.15, respectively. Furthermore, for the Travel and Adult datasets, Fleiss's Kappa yielded values of 0.18 and 0.17, respectively. Since all the values are positive, we conclude that the participants exhibited consistency in their evaluations, thus affirming the reliability of human evaluation.
    \paragraph{The Pearson correlation coefficient}
        As a supplementary description to the Table~\ref{tab:pearson}, we present a heatmap illustrating the correlation coefficients between human evaluations and automated assessment metrics.
        \begin{figure}[ht]
        \centering     \includegraphics[width=0.90\linewidth]{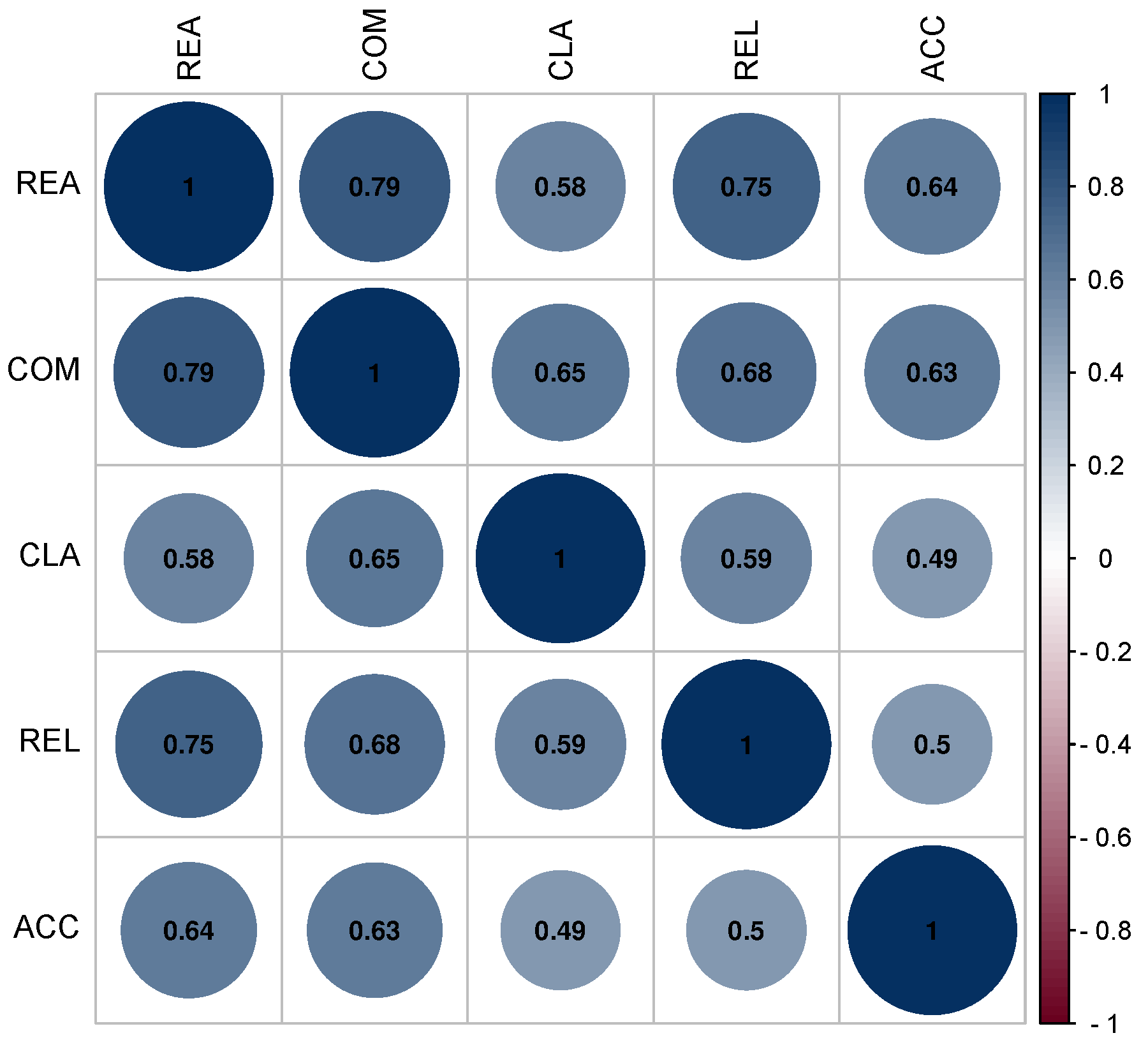}
            \caption{The Pearson correlation coefficient between manual and automatic metrics. "ACC." is the probability of a given sample being classified as reasonable by an SVM. The p-value of significance is 0.05.}
            \vspace{-5pt}
            \label{fig:pearson}
    \end{figure}
    \paragraph{Repetition rate.} It represents the proportion of duplicated instances in the generated samples compared to all generated samples. 
    A well-performing system should exhibit a lower degree of replication.

    \paragraph{Area Under the Curve (AUC).} AUC is the area under the receiver operating characteristic curve. 
    We use it to evaluate our model and the baseline to provide a comprehensive performance evaluation. 
 
     \begin{table*}[ht]
    \centering
    \resizebox{\textwidth}{!}{
    \begin{tabular}{lcccccccccccccc}
    \thickhline
    Dataset   & \multicolumn{4}{c}{Travel}                                                                          & \multicolumn{1}{c}{} & \multicolumn{4}{c}{Adult Income}                                                                    & \multicolumn{1}{c}{} & \multicolumn{4}{c}{HELOC}                                                                           \\ \cline{2-5} \cline{7-10} \cline{12-15} 
    Model     & \multicolumn{1}{c}{LR $\uparrow$} & \multicolumn{1}{c}{DT $\uparrow$} & \multicolumn{1}{c}{RF $\uparrow$} & \multicolumn{1}{c}{Mean $\uparrow$} & \multicolumn{1}{l}{} & \multicolumn{1}{c}{LR $\uparrow$} & \multicolumn{1}{c}{DT $\uparrow$} & \multicolumn{1}{c}{RF $\uparrow$} & \multicolumn{1}{c}{Mean $\uparrow$} & \multicolumn{1}{l}{} & \multicolumn{1}{c}{LR $\uparrow$} & \multicolumn{1}{c}{DT $\uparrow$} & \multicolumn{1}{c}{RF $\uparrow$} & \multicolumn{1}{c}{Mean $\uparrow$} \\ \hline
    Original  & 0.85                   & 0.87                   & 0.95                   & 0.89                     &                      & 0.58                   & 0.76                   & 0.91                   & 0.75                     &                      & 0.76                   & 0.62                   & 0.78                   & 0.72                     \\
    GPT-Neo + GReaT & \textbf{0.76}                   & 0.68                   & \textbf{0.72}                   & \textbf{0.72}                     &                      & 0.54                   & 0.70                   & 0.86                   & 0.70                     &                      & \textbf{0.66}                   & \textbf{0.60}                   & \textbf{0.78}                   & \textbf{0.68}                     \\
    GPT-2 + GReaT   & \underline{0.69}                   & 0.65                   & \underline{0.68}                   & 0.67                     &                      & 0.54                   & 0.70                   & 0.86                   & 0.70                     &                      & 0.54                   & 0.57                   & 0.68                   & 0.60                     \\
    GPT-Neo + Ours  & 0.61                   & \textbf{0.79}                   & 0.66                   & \underline{0.69}                     &                      & 0.87                   & 0.71                   & 0.87                   & 0.82                     &                      & \underline{0.64}                   & 0.57                   & \underline{0.77}                   & \underline{0.66}                     \\
    GPT-2 + Ours    & 0.62                   & \underline{0.73}                   & 0.63                   & 0.66                     &                      & 0.57                   & 0.68                   & 0.82                   & 0.69                     &                      & 0.54                   & 0.57                   & 0.72                   & 0.61                     \\ \thickhline
    \end{tabular}}
    \caption{AUC of classifiers trained with synthesized data using different models (listed on the left) on three datasets.}
    \label{tab:auc}
    \end{table*}
    
        In terms of the performance of the binary classifier (less than 2\%) trained on synthetic data, Table~\ref{tab:auc} illustrates that the data generated by our policy is similar to that of the baseline.  However, it is worth noting that our model substantially improved the AUC metric of LR by over 30\% on the Adult dataset, demonstrating the enhanced distributional consistency of our synthetic data, thereby enabling better applicability of the linear model.
        Furthermore, as the number of tabular features increased, the synthetic data generated by both methods tended to be more similar to the original data. It implies that language model-based table generation methods can exhibit advanced simulating capabilities when features are sufficient.

    \begin{table}[ht]
    \resizebox{\linewidth}{!}{
    \begin{tabular}{lccc}
    \thickhline
    Datasets     & \multicolumn{1}{l}{Travel}& \multicolumn{1}{l}{Adult Income} &  \multicolumn{1}{l}{HELOC}  \\ \hline
    Original     & 53.14\%    & 0.11\%                            & 5.61\%   \\
    Neo + GReaT  & 16.00\% & 0.01\%                             & 0.00\%   \\
    GPT-2 + GReaT      & 22.25\%  & 0.00\%                             & 0.00\%   \\
    GPT-Neo + Ours      & 16.00\%  & 0.01\%                             & 0.00\%    \\
    GPT-2 + Ours      & 22.67\%  & 0.00\%                             & 0.00\%   \\
    \thickhline
    \end{tabular}}
    \caption{Repetition rate (with top-p of 0.9 and temperature of 0.9 in decoding). }
    \label{tab:rep}
    \end{table}
    
        Considering the consistency of label distributions between synthetic and actual data, we first computed the KL divergence between the label distributions of synthetic data for testing and the original dataset. For the Adult dataset, the mean KL divergence was 0.01 $\pm$ 0.02 and a maximum value of 0.04. For the Travel dataset, the mean KL divergence was 0.02 $\pm$ 0.02 and a maximum value of 0.03. For the HELOC dataset, the mean KL divergence was 0.01 $\pm$ 0.01 and a maximum value of 0.03. Since the KL divergences between the label distributions of all synthetic datasets and the original dataset were less than 0.05, we conclude that the label distribution of the testing dataset is similar to that of the actual data.

        The results presented in Table~\ref{tab:rep} demonstrate that PPO does not significantly increase the replication rate (less than 1\%). We observed that for Adult Income and HELOC datasets, our method seldom generates results identical to those in the original dataset. 
        However, for the Travel dataset, given its mere seven features, with four having a binary domain, replication in the generation process unavoidably occurs.
 
 \section{Implementation Details}
 
     \begin{table*}[ht]
    \begin{tabular}{p{4.4cm}p{4.3cm}p{5.9cm}}
    \thickhline
    \multicolumn{1}{c}{\textbf{Travel Customers}} & \multicolumn{1}{c}{\textbf{Adult Income}} & \multicolumn{1}{c}{\textbf{HELOC}} \\ \hline
    Your task is to \textcolor{red}{Core Part}, referring to the following set of 3 positive customers who haven't churned from a travel company and 3 negative customers who have churned from a travel company: & Your task is to \textcolor{red}{Core Part}, referring to the following set of 3 positive adults who earn annual incomes which exceed \$50K and 3 negative adults who earn annual incomes which don't exceed \$50K: & Your task is to \textcolor{red}{Core Part}, referring to the following set of 3 positive individuals who have never been late for payments by more than 90 days over a period of 24 months since the account of Home Equity Line of Credit (HELOC) was opened and 3 negative individuals who have been late for payments at least 90 days by at least once over a period of 24 months since the account of Home Equity Line of Credit (HELOC) was opened:\\ 
    \makecell[l]{
    -{}-{}-\\
    POSITIVE CUSTOMERS\\
    $$[1]$$ FEATURES: ''\textcolor{brown}{Features} \\ \textcolor{brown}{of customer 1}''\\
    DESCRIPTION: ''\textcolor{brown}{Descri-} \\ \textcolor{brown}{ption of customer 1}''\\
    ...\\
    NEGATIVE CUSTOMERS\\
    $$[1]$$ FEATURES: ''\textcolor{brown}{Features} \\ \textcolor{brown}{of customer 4}''\\
    DESCRIPTION: ''\textcolor{brown}{Descri-} \\ \textcolor{brown}{ption of customer 4}''\\
    ...\\
    -{}-{}-\\
    Your task is to \textcolor{red}{Core Part}.} 
    & 
    \makecell[l]{
    -{}-{}-\\
    POSITIVE ADULTS\\
    $$[1]$$ FEATURES: ''\textcolor{brown}{Features} \\ \textcolor{brown}{of adult 1}''\\
    DESCRIPTION: ''\textcolor{brown}{Descri-} \\ \textcolor{brown}{ption of adult 1}''\\
    ...\\
    NEGATIVE ADULTS\\
    $$[1]$$ FEATURES: ''\textcolor{brown}{Features} \\ \textcolor{brown}{of adult 4}''\\
    DESCRIPTION: ''\textcolor{brown}{Descri-} \\ \textcolor{brown}{ption of adult 4}''\\
    ...\\
    -{}-{}-\\
    Your task is to \textcolor{red}{Core Part}.} 
    & 
    \makecell[l]{
    -{}-{}-\\
    POSITIVE INDIVIDUALS\\
    $$[1]$$ FEATURES: ''\textcolor{brown}{Features of individ-} \\ \textcolor{brown}{ual 1}''\\
    DESCRIPTION: ''\textcolor{brown}{Description of indi-} \\ \textcolor{brown}{vidual 1}''\\
    ...\\
    NEGATIVE INDIVIDUALS\\
    $$[1]$$ FEATURES: ''\textcolor{brown}{Features of individ-} \\ \textcolor{brown}{ual 4}''\\
    DESCRIPTION: ''\textcolor{brown}{Description of indi-} \\ \textcolor{brown}{vidual 4}''\\
    ...\\
    -{}-{}-\\
    Your task is to \textcolor{red}{Core Part}.}  \\ \hline
    
    \textcolor{red}{Core Part}: explain the reason why the customer with the FEATURES: ''\textcolor{brown}{Features}'' and DESCRIPTION: ''\textcolor{brown}{Description}'' \textcolor{orange}{has} \textcolor{orange}{(hasn't)} churned from the travel company
    & 
    \textcolor{red}{Core Part}: explain the reason why the adult with the FEATURES: ''\textcolor{brown}{Features}'' and DESCRIPTION: ''\textcolor{brown}{Description}'' earns an annual income which \textcolor{orange}{doesn't exceed} \textcolor{orange}{(exceeds)} \$50K 
    & 
    \textcolor{red}{Core Part}: explain the reason why the individual with the FEATURES: ''\textcolor{brown}{Features}'' and DESCRIPTION: ''\textcolor{brown}{Description}'' has \textcolor{orange}{been late for payments at least 90 days by at least once} \textcolor{orange}{(never been late for payments by more than 90 days)} over a period of 24 months since the account of Home Equity Line of Credit (HELOC) was opened \\ \thickhline
    \end{tabular}
    \caption{Prompts used to generate explanations. \textcolor{red}{Red text} represents the core part of a prompt; \textcolor{brown}{brown text} represents the features or description of a specific sample; \textcolor{orange}{orange text} represents an alternate text determined by the target value of a sample.}
    \label{tab:prompt}
    \end{table*}
     
    \begin{table}[H]
    \centering
    \begin{tabular}{lcc}
    \thickhline
    Device           & Server A                      & Server B               \\ \hline
    Memory           & 13G                           & 1012G                  \\
    CPU              & Intel Core & AMD \\
    & i5-1135G7 & EPYC 7763\\
    & 2.40G Hz & 2.45G Hz \\
    GPU              & NVIDIA & NVIDIA\\
    &Tesla T4 16G & A100 80G\\
    Operating & Ubuntu &Ubuntu \\
    System &18.04.6 LTS    & 20.04.6 LTS     \\ \thickhline
    \end{tabular}
    \caption{Experimental hardware environment.}
	\label{tab: Hardware}
    \end{table} 
    For the experimental setup, we utilized two servers. All work related to the significant language model mentioned in our paper was conducted on Server B. Details of the experimental environment are provided in Table~\ref{tab: Hardware}.
    
    We employed the standard GPT-2~\citep{Radford2019LanguageMA} with 1.5 billion parameters and GPT-Neo~\citep{Black2021GPTneoLS} with 1.3 billion parameters models as generators and a self-attention-based Convolutional Neural Network (CNN)~\citep{726791}. 
    The CNN has 32 convolutional kernels, eight attention heads, and 512 neurons in the attention layer~\citep{NIPS2017_3f5ee243}. 

    For the interpretability task, we randomly selected 100 samples from the test set for explanation and human evaluation. 
    The LLM and interpreter used for describing samples were Vicuna-v1.5~\citep{zheng2023judging}, fine-tuned on LLaMA 2~\citep{meta2023introducing}. The model used for computing semantic similarity between descriptive texts was Sentence-Bert~\citep{reimers-gurevych-2019-sentence}.
        
    During training, we utilized the optimizer of AdamW~\citep{loshchilov2018fixing} with a learning rate of 1e-4 and a batch size of 16.
    During inference, the maximum generation length was 300, top-p~\citep{Holtzman2020The} was 0.9, temperature was 0.7, and repetition penalty coefficient was 1.2.
    
    The set of prompts we used for explanations is presented in Table~\ref{tab:prompt}.
        
    Regarding the questionnaires used, we posed the following four statements for each sample evaluated:
    
    (reasonableness) \textit{The model explanation exhibits minimal logical errors and represents rational interpretations of the question.}
    
    (relevance) \textit{The model explanation does not include irrelevant content to the topic.}
    
    (clarity) \textit{The model explanation is easy for me to understand.}
    
    (comprehensiveness) \textit{The model explanation covers all important features as I would expect.}
    
    To ensure the effectiveness of the feedback collected, we established a gold question and provided two options:
    
    (gold question) \textit{Before getting started, let's make sure that you understand the task. Please choose the task that you will do:}
    
    \textbf{A.} \textit{I will rate the quality of the decision provided by the model.}
    
    \textbf{B.} \textit{I will rate the quality of the explanation provided by the model.}
    
    We will only consider feedback from respondents who answered the gold question correctly with option A. Any feedback from respondents who omitted answers will also be deemed invalid.
    
    Furthermore, we randomly selected a sample from the dataset, placing it at the beginning of the questionnaire as a warm-up example to ensure the respondents understood the task.
    
\section{Additional Case Study}
\label{sec:appendix-casestudy}
    To specifically illustrate the effectiveness of our explanations, we conducted a case study by randomly selecting samples from the three datasets. 

    The explanations generated are presented in Table~\ref{tab:case_travel}, Table~\ref{tab:case_adult} and Table~\ref{tab:case_heloc}. We observed that our model precisely captured the underlying reasons that could lead to specific features and provided further implications based on these reasons. For instance, the model realized that booking a hotel room is a positive signal for customer retention. At the same time, low activity on social media may be a negative signal, while other features, such as age, may not significantly impact the retention of the target person. Additionally, through comparisons with other samples, the model recognized the significance of specific feature values. For instance, it noted that working 35 hours per week falls under the category of long working hours, and having an educational duration of 13 years is higher than the average level.

\begin{table*}[ht]
\begin{tabular}{p{1.5cm} p{2cm} p{3.5cm} p{9.5cm}}
\thickhline
\multicolumn{2}{l}{\multirow{2}{*}{Adult}} & Synthetic data         & age:54, workclass:Local-gov, fnlwgt:176977, education:Bachelors, educational-num:13, marital-status:Married-civ-spouse, occupation:Prof-specialty, relationship:Husband, race:White, gender:Male, capital-gain:0, capital-loss:0, hours-per-week:35, native-country:United-States, income:>50K \\
\multicolumn{2}{l}{}                                & Description generated  & John Doe is a 54 year old male from the United States who works as a professional specialist for local government. He has a Bachelor's degree and 13 years of formal education under his belt. John is married to his civilian spouse and is white in terms of both race and ethnicity. His weekly working hours are 35, with an annual income exceeding \$50k. \\
\multicolumn{2}{l}{User}                            & User's question       & Explain the reason why the adult has an income over \$50K. \\ \hline
\multicolumn{2}{l}{Ours} & Explanation generated & The reason he earns an annual income which exceeds \$50K is because \textcolor{blue}{he holds a position as a professional specialist} for \textcolor{blue}{local government}, where his expertise and experience likely command higher wages compared to other occupations within the same work class. Additionally, his completion of a \textcolor{blue}{Bachelor's degree and 13 years of formal education} contribute to his knowledge base and skillset, making him more valuable to his employer. Furthermore, \textcolor{blue}{John's marriage to someone outside of the military may provide additional support and resources}, helping him maintain a comfortable standard of living despite his relatively \textcolor{blue}{long working hours each week (35)}. Lastly, his status as \textcolor{blue}{a white male in the United States could be advantageous when negotiating salaries and benefits.} Overall, these factors combine to give John Doe access to better paying jobs than those without similar qualifications, resulting in an annual income above \$50K.  \\ \thickhline
\end{tabular}
\caption{Case Study of the Adult Income Dataset. \textcolor{blue}{Blue text} represents key-points in explanations.}
\label{tab:case_adult}
\end{table*}

\begin{table*}[ht]
\begin{tabular}{p{1.5cm} p{2cm} p{3.5cm} p{9.5cm}}
\thickhline
\multicolumn{2}{l}{\multirow{2}{*}{HELOC}} & Synthetic data         & RiskPerformance:Good, ExternalRiskEstimate:69, MSinceOldestTradeOpen:184, MSinceMostRecentTradeOpen:5, AverageMInFile:74, NumSatisfactoryTrades:31, NumTrades60Ever2DerogPubRec:0, NumTrades90Ever2DerogPubRec:0, PercentTradesNeverDelq:100, MSinceMostRecentDelq:-7, MaxDelq2PublicRecLast12M:7, MaxDelqEver:8, NumTotalTrades:31, NumTradesOpeninLast12M:0, PercentInstallTrades:23, MSinceMostRecentInqexcl7days:-7, NumInqLast6M:0, NumInqLast6Mexcl7days:0, NetFractionRevolvingBurden:3, NetFractionInstallBurden:-8, NumRevolvingTradesWBalance:1, NumInstallTradesWBalance:-8, NumBank2NatlTradesWHighUtilization:0, PercentTradesWBalance:33 \\
\multicolumn{2}{l}{}                                & Description generated  & With an external risk estimate of 69 and a minimum file score of 74, it appears that this HELOC may pose some level of risk but is still within acceptable parameters. Additionally, there have been no trades delinquent by more than 60 days or those ever reported as public records in the last two years. Furthermore, only one trade is considered \''derogatory,\'' indicating that the borrower's credit history is generally positive. It's worth noting that this HELOC has had zero trades open within the past year, which could suggest a low utilization rate. However, the maximum delinquency recorded over the past twelve months was seven, suggesting potential room for improvement.  \\
\multicolumn{2}{l}{User}                            & User's question        & Explain the reason why the applicant will not have a 90-day or more severe delinquency after opening a credit account. \\ \hline
\multicolumn{2}{l}{Ours}                            & Explanation generated & The explanation lies in the fact that the individual with the given features has shown consistent and reliable repayment behavior throughout their credit history. Despite being delinquent in the past, \textcolor{blue}{they have since made up for it by demonstrating perfect payment history in the subsequent periods.} Moreover, \textcolor{blue}{their current credit score} indicates that they are likely to continue making timely payments going forward. As a result, despite the presence of certain risk factors like high revolving burden and negative install burden, their overall creditworthiness is deemed sufficient to justify \textcolor{blue}{the assigned risk estimate of 69.}  \\ \thickhline
\end{tabular}
\caption{Case Study of the HELOC Dataset. \textcolor{blue}{Blue text} represents key-points in explanations.}
\label{tab:case_heloc}
\end{table*}

\end{document}